%% file: main.tex
\documentclass[conference]{IEEEtran}
\IEEEoverridecommandlockouts
\usepackage{cite}
\usepackage{amsmath,amssymb,amsfonts}
\usepackage{graphicx}
\usepackage{textcomp}
\usepackage{xcolor}
\def\BibTeX{{\rm B\kern-.05em{\sc i\kern-.025em b}\kern-.08em
    T\kern-.1667em\lower.7ex\hbox{E}\kern-.125emX}}

\usepackage{makecell}
\usepackage{algorithm}
\usepackage{algorithmicx}
\usepackage{algpseudocode}
\usepackage{enumerate}
\usepackage{multicol}
\usepackage{multirow}
\usepackage{todonotes}
\usepackage{authblk}
\newtheorem{example}{Example}

\begin{document}

\title{Knowledge Graph Extraction from Videos}
%%\thanks{Supported by organization x.}}
%
%\titlerunning{Deep Learning for Logical Annotation and its Application to Videos}
\author[1]{Louis Mahon}
\author[1]{Eleonora Giunchiglia}
\author[1]{Bowen Li}
\author[1]{Thomas Lukasiewicz}
\affil[1]{Computer Science Department, Oxford University}
%
%\authorrunning{} %% F. Author et al.}
\maketitle              % typeset the header of the contribution
\begin{abstract}
\input{text/abstract.tex}
\end{abstract}

\input{text/intro.tex}

\input{text/datasets.tex}

\input{text/method.tex}

\input{text/results.tex}

\input{text/discussion}

\input{text/conclusion.tex}

\bibliographystyle{IEEEtran}
\bibliography{bibliography}

\clearpage
\input{text/appendix}
\end{document}

%% file: text/abstract.tex
Nearly all existing techniques for automated video annotation (or captioning) describe videos using natural language sentences. However, this has several shortcomings: (i) it is very hard to then further use 
the generated natural language annotations in automated 
%not possible 
 data processing, (ii) generating natural language annotations requires to solve  the hard subtask of generating semantically precise and syntactically correct natural language sentences, which is actually unrelated to the task of video annotation, (iii) it is difficult to quantitatively measure performance, as standard metrics (e.g., accuracy and F1-score) are inapplicable, and (iv)~annotations are language-specific. In this paper, we propose the new task of knowledge graph extraction from videos, i.e., producing a description in the form of a knowledge graph of the contents of a given video. Since no datasets exist for this task, we also include a method to automatically generate them, starting from  datasets where videos are annotated with natural language. We then describe an initial deep-learning model for knowledge graph extraction from videos, and report results on MSVD* and MSR-VTT*, two datasets obtained from MSVD and MSR-VTT using our method.

%% file: text/intro.tex
\section{Introduction}

Recent progress in deep learning shows exciting potential for visual understanding. 
%Driven by this progress, and by important use cases (such as autonomous driving, human-robot interaction, video captioning, and video search/recommendation), deep learning has produced 
Promising results have been produced in the tasks of video annotation (or captioning) using natural language (see, e.g., \cite{zhang2019,gao2017,zhou2018}) and video question answering (see, e.g., \cite{zhao2017,zeng2017,lei2018}).

\begin{figure*}[t]
\begin{tabular}{c l c}
    \multirow{13}{*}{\includegraphics[width=0.32\textwidth]{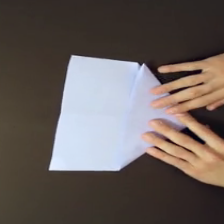}} & 
    \textbf{(1) MSR-VTT ground truth:} & \multicolumn{1}{l}{\textbf{(7) Visual representation of the}}    \\
    &\text{a person folds a piece of white paper} & \multicolumn{1}{l}{\textbf{predicted knowledge graph:}} \\[1ex]
    &\textbf{(2) Predicted caption from \cite{olivastri2019end}:} &\multirow{8}{*}{\includegraphics[width=0.283\textwidth]{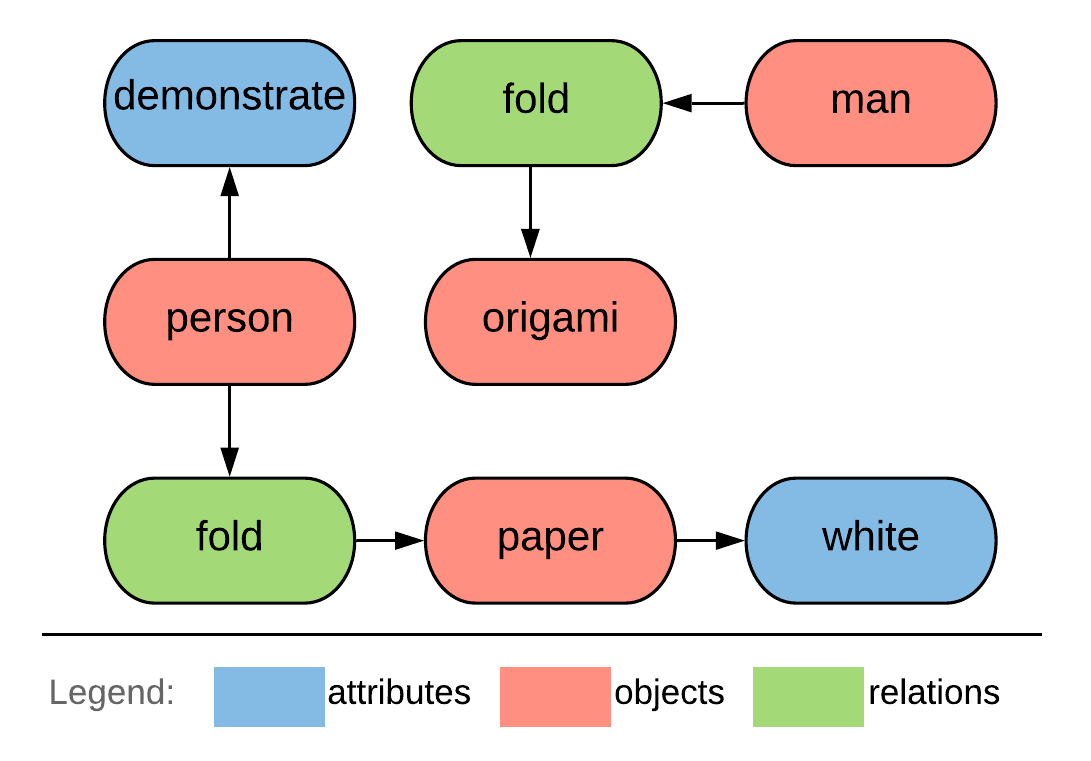}} \\ 
    &\text{a person is folding a piece of paper}\\
    & \vspace*{-1.1ex} \\
    \cline{2-2}
    &  \vspace*{-0.6ex}\\
    &\textbf{(3) MSR-VTT* ground truth:} \\
    &\textit{fold}(\textit{person, piece}) \\[1ex]
    &\textbf{(4) Predicted individuals:} \\ 
    &\textit{person, man, paper, origami} \\
    &\textbf{(5) Predicted facts:} \\
    &\multicolumn{2}{l}{\textit{demonstrate}(\textit{person}), \textit{fold}(\textit{person, paper}) 
    \textit{white}(\textit{paper}), \textit{fold}(\textit{man, origami})}\\
    &\textbf{(6) Some inferred facts:} \\
    &\multicolumn{2}{l}{\textit{change}(\textit{person,\,paper}), \textit{fold}(\textit{male,\,origami})} 
    \end{tabular}
   
    \medskip 
    \caption{%
        The first frame from a video in MSR-VTT*, with (1) the ground-truth natural language annotation in MSR-VTT, (2) the natural language annotation  produced by \cite{olivastri2019end}, 
        (3)~the ground-truth set of facts in MSR-VTT*, 
        (4) the individuals predicted by our system, (5) the facts predicted by our system, and (6)~some further inferences that can be made on top of these facts.
   \vspace*{-2ex} }
    \label{fig:msrvtt-ex}
\end{figure*}

However, annotations written in natural language (NL) are not ideal, for several reasons. %\textcolor{red}{Firstly, it is difficult to automatically extract and process information expressed in NL. 
Firstly, it is difficult to use NL sentences in subsequent automated data processing tasks.
One example of such data processing is database search. If we have a database of videos annotated with NL and want to search for all videos depicting a man, we cannot simply perform string matching on the annotations. This would, e.g., produce a false positive for the annotation ``sailors man the ship'', and a false negative for ``a firefighter puts on his coat''. Another example of automated data processing is logical inference. For example, if an NL annotation model correctly identifies that there is a man throwing a ball in an image, it cannot then conclude that there is a male throwing a ball in the image. That is,  a logical inference rule of the form  $\forall x~\mathit{man}(x) \Rightarrow \mathit{male}(x)$ cannot easily be applied. Automated inference in NL is known to be a difficult problem \cite{bowman2015large,belinkov2019don}. Again, string matching would not provide a solution, as the above examples attest. A~second drawback of NL annotations is that they require more work than just understanding the contents to be annotated. Traditionally, text generation has been divided into a semantic stage of determining what to say, and then a realization stage of determining how to say it \cite{mckeown1985discourse, levelt1993speaking}. The former corresponds to understanding the contents being annotated and is the task that we are interested in, but deep learning NL annotation models must, at the same time, learn to perform the latter, which means learning the language's complex meaning and grammatical structure. This is irrelevant to the goal of annotation and so needlessly makes the task more difficult. Thirdly, it is not easy to interpret and quantitatively measure the performance of NL annotations. The agreement between a ground-truth and the predicted sentence cannot be measured by standard metrics, such as accuracy and F1-score. Instead, they require specially designed ones (e.g., BLEU \cite{papineni2002bleu}, METEOR \cite{banerjee2005meteor}, and LEPOR \cite{han2012lepor}). Recognizing the imperfection of each of these, results typically report scores on multiple metrics, none of which have a simple and intuitive interpretation. This makes it difficult to evaluate how well a model is performing. A fourth problem is that NL annotations are specific to a single language. A model trained to produce annotations in English cannot produce annotations in German, and it is non-trivial to translate the English annotations into German ones. This is a particular drawback for low-resource languages. 

Replacing NL annotations with knowledge graphs avoids all the above problems. A knowledge graph (KG) specifies the {\sl individuals} (i.e., the objects) present in the video and the facts that hold true of these individuals. A {\sl fact} expresses a relation between two individuals (e.g., $\mathit{fold(person, paper)}$) or an attribute of an individual (e.g., $\mathit{white(paper)}$). A visual representation of a KG is given in Fig.~1, where nodes representing individuals are depicted in red, nodes representing attributes are depicted in blue, and nodes representing relations are depicted in green.
KGs are machine-readable, so enable automated data processing, such as searching for all videos depicting a man or applying the inference rule $\forall x~\mathit{man}(x) \Rightarrow \mathit{male}(x)$. They are determined only by the semantics of the video contents, and do not involve a semantically and syntactically complex natural language structure. They can be evaluated using standard machine-lear\-ning metrics, such as accuracy and F1-score, which makes it easy to interpret performance. For each data point, the ground truth and the prediction are sets of facts, so, e.g., the F1-score can be computed by counting the facts that appear in both as the true positives, those that appear in the former only as false negatives, etc. Finally, each fact, and hence each KG, can be trivially translated by translating one component at a time. For example, we can translate $\mathit{white(paper)}$ to German by translating $\mathit{white}$ to $\mathit{weiss}$ and $\mathit{paper}$ to $\mathit{Papier}$, giving $\mathit{weiss(Papier)}$. (As discussed in Sections \ref{sec-dataset} and \ref{sec-discussion}, the components of each fact denote particular senses of words, so word-sense disambiguation is not an issue.) For these reasons, we propose to produce annotations using KGs rather than NL. 

These advantages have begun to be recognized in neighbouring areas. The entire task of open information extraction is motivated by use cases that can only be met by a structured representation of information, and not by the same information expressed in natural language. In computer vision, researchers have started to argue for the importance of basic visual reasoning (see, e.g.,~\cite{johnson2016}) and the ability to leverage external knowledge (see, e.g.,~\cite{wang2016}). A number of works aim to extract a structured description from an input image to reason about the identified objects (see, e.g., \cite{wang2017}). In the video understanding field, there have been only preliminary attempts~\cite{Vasile}, though, recently, \cite{curtis2020hlvu} has sketched some planned future work to manually create a dataset of videos annotated with facts.

In this paper, we describe a method for automatically generating datasets of videos and corresponding KGs. The method uses a rule-based semantic parser, based on the Stanford~parser \cite{qi2018universal}. The generated datasets are sufficient for the development of KG extraction models. Additionally, we introduce one such model, trained and tested on two generated datasets, which achieves a superior performance to \cite{Vasile}, the only  existing work to attempt the same task (F1-score of $14.0$ vs.\ $6.1$). We also compare to an artificial baseline, conduct ablation experiments that demonstrate the efficacy of each of the model's components, and present qualitative results. 
This model enjoys all the advantages described above: it facilitates automated data processing, it does not require modelling complex NL syntax and semantics, its performance is easy to evaluate, and the resulting annotations are easy to translate. 

\smallskip 
\begin{example}
Fig.~\ref{fig:msrvtt-ex} shows an example where the input is a video of someone folding paper. Given this video, our mo\-del detects the contained individuals, namely, \textit{person, man, paper,} and \textit{origami}. Then, it predicts which facts are true of these individuals: \textit{demonstrate}(\textit{person}), \textit{fold}(\textit{person, paper}), \textit{white}(\textit{paper}), and \textit{fold}(\textit{man, origami}). As an example of automated data processing, it may then exploit an ontology to predict more facts that are true for the video, e.g., the rule~$\forall x,y$ $(\textit{fold}(x,y) \,{\Rightarrow}\,\textit{change}(x,y))$ to infer the fact \textit{change}(\textit{person, paper}). The corresponding KG is shown in~(7).~\hfill$\lhd$
\end{example} 

\smallskip 
The main contributions of this paper are briefly as follows. 
\begin{itemize}
    \item We propose the task of extracting KGs from videos. We provide arguments for the superiority of this approach to NL annotations, and support these arguments empirically. 
    \item We describe a method for automatically generating datasets of videos and corresponding KGs, and describe its application to generate two datasets. %These datasets, along with the code for the generation method, will be made freely available on publication.
    \item We introduce a model for KG extraction, trained and tested on the above two datasets. This outperforms the existing work to attempt the same task. We also compare to an artificial baseline and conduct ablation studies on the main components of the model. 
\end{itemize}

The rest of this paper is organized as follows. In Section~\ref{sec-dataset}, we present our method for automatically generating datasets. In Section \ref{sec-method}, we describe our KG extraction model, and then report experimental results in Section \ref{sec-results}. Section \ref{sec-discussion} provides a further discussion, and Section~\ref{sec-conclusion} summarizes our main contributions and gives an outlook on future work.

%% file: text/datasets.tex
\section{Dataset Generation} \label{sec-dataset}

In the absence of appropriate datasets, we devise an automated method to generate them. There are several NL-an\-no\-ta\-ted datasets, which we refer to as video-captioning datasets. Our method begins with these datasets and converts the NL annotations into sets of facts (which are equivalent to KGs). We now describe this method with reference to its application to two well-known video-captioning datasets: MSVD \cite{chen2011collecting} and MSR-VTT \cite{xu2016msr}. We denote the generated datasets MSVD* and MSR-VTT*, respectively, and Section~\ref{sec-results} reports the results of our proposed model on these datasets. 
%The dataset generation code  will be made freely available on publication. % at $<$insert-git-url$>$. 

In both MSVD and MSR-VTT, the training examples are composed of a video and a NL sentence describing the contents of that video. We parse these sentences using a rule-based parser based on the Stanford NLP syntactic parser \cite{qi2018universal}. This produces a dependency parse of the sentence, which identifies the part of speech for each word, and the syntactic relations that hold between them. We then apply a sequence of rules to extract the semantic information from the sentence, and use it to form a KG, which is composed of a set of facts. Each fact contains a predicate and the corresponding arguments: $\langle subject, predicate, object\rangle$ triples in the case of binary predicates, and $\langle subject, predicate \rangle$ pairs in the case of unary predicates (see Figs. \ref{fig:msrvtt-ex} and \ref{fig:msvd-msrvtt-ex}). A full description of this parser can be found in the appendix. 

The next step is to link all predicates, and individuals to entities in an ontology. The ontology that we use here is WordNet \cite{miller1990introduction}. The linking method is context-sensitive, i.e., for a given word, it finds the most suitable WordNet synset using the surrounding words in the NL annotation. This is done by comparing the word vectors for the surrounding words to the word vectors for the synset definition. Again, full details are given in the appendix. Linking to an ontology formalizes the vocabulary and allows the application of inference rules to augment the information produced directly by our annotation system. It also means that the components of our KGs correspond to particular senses of words, rather than the words themselves. For example, ``bank'' referring to the financial institution, and ``bank'' referring to the side of a river would be linked to different WordNet synsets, and so would appear as different items in a KG. 

Words that appear only a small number of times across the dataset would be difficult to reach a good performance on. For this reason, we exclude all words that appear fewer than 50 times and also some common semantically weak verbs, as~detailed in the appendix.

The method described so far produces only non-negated facts, from here on referred to as $T$. This means a model could learn to simply predict every potential fact to be true. To prevent this, we use the local closed-world assumption \cite{dong2014knowledge} to create a set of negated facts, from here on referred to as~$F$, which the model must also learn to predict. For each fact in the description, we obtain a corrupted version by replacing the predicate with a different predicate from the vocabulary. For example, the fact $fold(person,paper)$, which appears in Fig. \ref{fig:msrvtt-ex}, could be corrupted to $\neg throw(person,paper)$. This negated fact is then added to $F$. 

As a final step, we merge all KGs for each video by taking the union of the facts of their parses. In MSVD and \mbox{MSR-VTT}, each video appears with multiple captions, and each produces a separate training example. In MSVD* and MSR-VTT*, each video appears in only one training example. After exclusion and merging, the numbers of distinct individuals, distinct predicates, and data points with at least one fact or negated fact are as reported in Table~\ref{tab:msvd-results}.
\begin{table*}[t]
\centering
\caption{\small Numbers of distinct predicates and individuals that were included in our final datasets. Predicates and individuals were included if they appeared at least 50 times. The final column shows the number of data points that remained with non-empty captions even after infrequently occurring words were excluded.}
    \label{tab:msvd-results}
    \medskip 
\scalebox{0.92}{
\begin{tabular}{c|c|c|c|c|c|c|} 
\cline{2-7}
%& \thead{Num Training Examples} & \thead{Num Individuals} & \thead{Num Unary Predicates} & \thead{Num Binary Predicates} & \thead{Num Predicates} & \thead{Num Non-empty Training Examples}  \\
& \thead{Num Training Examples} & \thead{Num Individuals} & \thead{Num Attributes} & \thead{Num Relations} & \thead{Num Facts} & \thead{Num Non-empty Training Examples} \\ 
\hline
\multicolumn{1}{|c|}{\textbf{MSVD*}} & 1970 & 122 & 48 & 69 & 117 & 1800\\ 
 \hline
\multicolumn{1}{|c|}{\textbf{MSR-VTT*}} & 10000 & 372 & 235 & 113 & 348 & 9802 \\
  \hline
 \end{tabular}\vspace*{-2ex}
 }
\end{table*}

%% file: text/method.tex
\section{Our Model} \label{sec-method}
In this section, we describe a deep learning model (illustrated in Fig.~\ref{fig:network}) for extracting KGs from videos. We first describe a general method for extracting a KG from an arbitrary input, and then detail how this method applies to the case where the inputs are videos.

\subsection{Knowledge Graph Extraction Model}

Given a set of possible inputs $X$ ({videos in our case)} and a knowledge base $K\!B$ (of ontological domain knowledge about~$X$), let our vocabulary consist of the set $P$ of all predicates that appear in $K\!B$  and the set~$A$ of all individuals that appear in $K\!B$. Then, the neural architecture for KG extraction consists of the following four components:  
\begin{enumerate}
    \item an encoder $f\colon X \rightarrow Z$, 
    \item a multi-classifier $g
\colon Z \rightarrow (0,1)^{|A|}$,
    \item a set of predicate multilayer perceptrons (MLPs): $\{m_p \mid p \in P\}$,
    \item a set of trainable individual vectors: $\{v_a \mid a \in A\}$, 
\end{enumerate} 
where $Z$ is the space of extracted feature vectors.

The data for training the above neural architecture consists of 4-tuples $(x,c,T,F)$, where $x \in X$ is the input to be annotated (a 4D video tensor in our case), $c \subseteq A$ is the set of individuals that are present in the input, and $T$ (resp., $F$) is a set of facts (resp., negated facts) containing these individuals.

\smallskip 
\begin{example}
Consider again the video in Fig.~\ref{fig:msrvtt-ex}. 
The training tuple associated with the video is 
\[
\left(x, \{person, piece\}, \{fold(person, piece)\}, F\right) \,,
\] 
where $x$ represents the video itself, and $F$ is a set of negated facts not containing \mbox{$\neg$\textit{fold$($person, piece$)$}}. An example of a negated fact that belongs to $F$ is $\neg$\textit{throw$($person, piece$)$}.\hfill$\lhd$
\end{example}

\smallskip 
During training, we first compute the feature vector \mbox{$f(x) = e \in Z$}. Then, we feed this to the multi-classifier to predict the individuals present in $x$, $\hat{c} = g(e)$, and compute a binary cross-entropy loss (BCE) for each individual in the vocabulary. This gives
\[
\mbox{$\mathcal{L}_{c} = \sum_{j=0}^{|A|} \mathcal{L}_{bin}(\hat{c}_j,c_j) \,,$}
\]
where $c_j\in \{0,1\}$ and $\hat{c}_j  \in (0,1)$ are the ground truth and prediction as to whether the $j$th individual in the vocabulary is present, respectively, and $\mathcal{L}_{bin}$ represents BCE. 
We also produce predictions for the facts and negated facts in $T$ and $F$ by selecting the individual vectors corresponding to the individuals in the ground truth, and feeding these vectors (along with the video encoding) to the predicate MLPs corresponding to the predicates in the ground truth. 
That is, for each fact $t \in T$
with $t = p(a,b)$,
we compute a prediction $\hat{t} = m_{p}(e,v_{a},v_{b})$, and similarly for negated facts. A naive brute-force method would be to run every predicate MLP on every individual vector. This would require $|P| \times |A|$ forward passes for every video. Selecting only the present individuals reduces this to $|P| \times n$, where $n$ is the number of individuals in the input video, which is typically less than three.

The vectors $v_a$ and $v_b$ are initialized to the corresponding word2vec word vectors \cite{le2014distributed}. For example, $v_{man}$ is initialized to the word2vec vector for the word ``man''. These vectors are updated during training along with the network weights. 

The training data contains many more negated facts than facts. To counteract this imbalance, we weigh the loss for facts and negated facts inversely to how many there are. The prediction loss is again calculated using BCE, this time applied to each fact (resp., negated fact). Formally, 
\[
\mbox{$\mathcal{L}_{p} = - \frac{1}{2|T|}\sum_{t \in T} \log(\tilde{t}) - \frac{1}{2|F|}\sum_{f \in F} \log(1-\tilde{f}) \,.$}
\]
The total backpropagated loss is then a simple summation:
\[
\mathcal{L} = \mathcal{L}_{c} + \mathcal{L}_{p} \,.
\]

At inference time, given an input $x$, we first compute the feature vector (step (i)), then threshold the outputs of the multi-classifier to obtain the set $I$ of predicted individuals (step (ii)):
\[
\hat{I} = \{a \in A \mid  g(f(x))_a > 0.5\} \,.
\]
As we no longer have access to the ground truth set of individuals, we instead select the \emph{predicted} individuals and pass them to the predicate-MLPs (step (iii)). This again avoids having to run all predicate-MLPs on all individuals. Specifically, to predict the unary relations that hold of these individuals, we run all predicate-MLPs in our vocabulary on all individual vectors corresponding to individuals in~$\hat{I}$, and to predict the binary relations, we do the same on all pairs of individuals in~$\hat{I}$. The total set $\hat{T}$ of predicted facts is the union of the two:
\begin{align*}
  &  U = \{p(a) \mid p \in P, a \in \hat{I}, m_p(a) > 0.5\},  \\
 &   B = \{p(a,b)\mid p \in P, a, b \in \hat{I}, m_p(a,b) > 0.5\},  \\
&    \hat{T} = U \cup B.
\end{align*}
Now, as an example of automated information processing, we can apply the additional knowledge from $K\!B$ to the predicted facts~$\hat{T}$ to obtain an additional set of inferred facts $T'$:
\begin{align*}
& U' = \{p(a)\mid p \in P, a \in \hat{I},  K\!B\cup \hat{T}\models p(a)\}, \\
& B' = \{p(a,b)\mid  p \in P, a,b \in \hat{I}, K\!B\cup \hat{T}\models p(a,b)\}, \\
& T' = U' \cup B'. 
\end{align*}
The annotation is then updated with these inferences\,(step\,(iv)):
\[
\hat{T} \gets \hat{T} \cup T'.
\]

\begin{figure*}
    \centering
    \includegraphics[height=8cm]{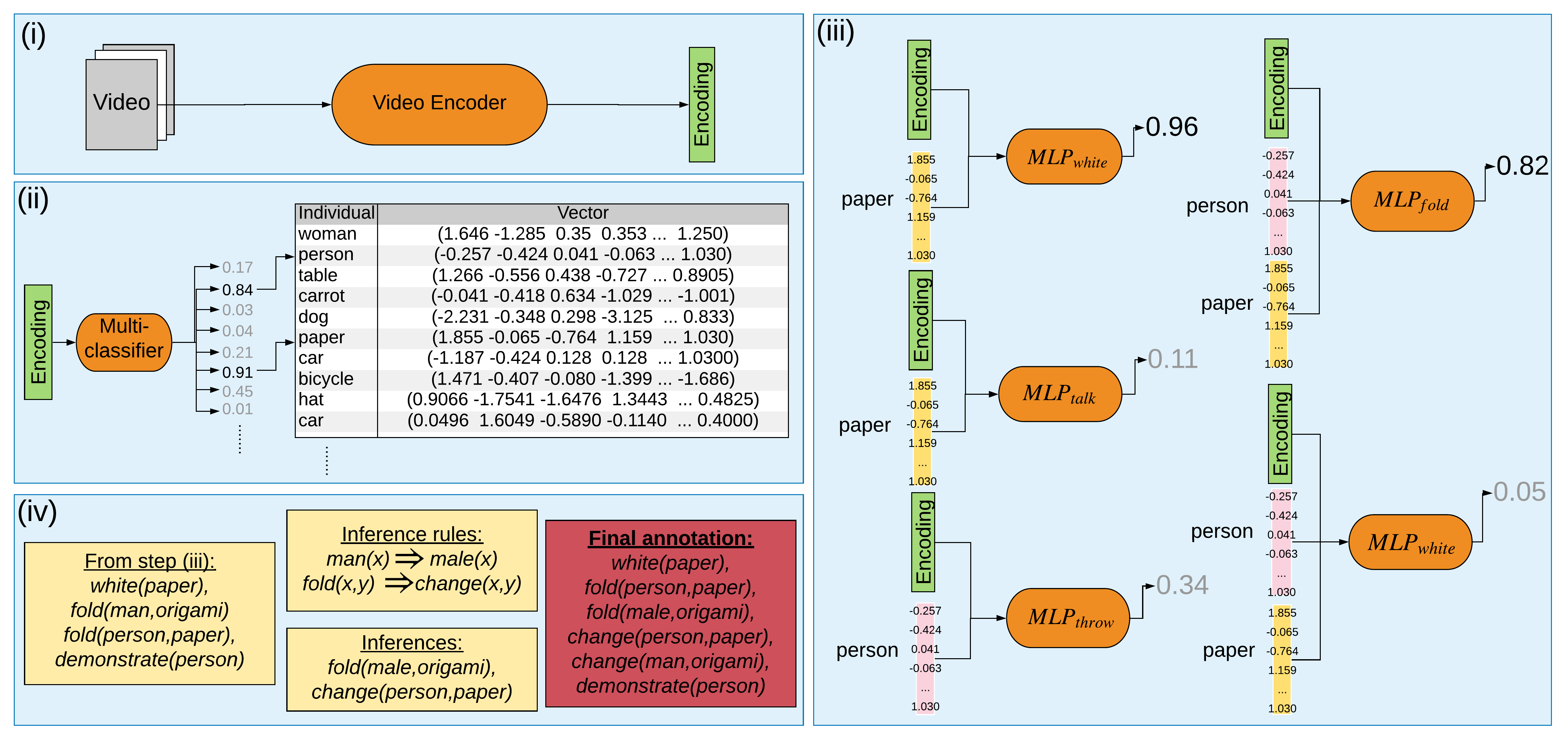} \\
    \caption{Depiction of how the neural architecture processes a video of a man folding a piece of paper (as in Fig.~\ref{fig:msrvtt-ex}). The four stages are: (i) encode the video, (ii) predict individuals, (iii) predict facts, and  (iv) add facts inferred from $K\!B$.
    }
    \label{fig:network}
\end{figure*}

\begin{example}
Consider again the video in Fig.~\ref{fig:msrvtt-ex}, and the schema of the model in Fig.~\ref{fig:network}.
Following the steps in Fig.~\ref{fig:network}, the KG for the video in Fig.~\ref{fig:msrvtt-ex} is produced in four steps. 
\begin{enumerate}[(i)]
    \item The input video $x$ is mapped into a fixed-length feature vector $e$ by the encoder $f$.
    \item The encoding $e$ is fed into a multi-classifier MLP $g$, which returns the probability that each individual in the vocabulary is present in $x$. The probabilities are then thresholded at $0.5$, and we obtain the set \[I = \{person, man, paper, origami\}.\]
    \item Consider the individual $paper$. The encoding $e$, together with the vector $v_{paper}$ is passed through all the MLPs, and a fact is returned only when the prediction of the MLP is greater than 0.5, such as $white(paper)$. The same procedure is then repeated for each predicted individual, and each ordered pair of predicted individuals, producing 
    \begin{gather*}
    \hat{T} = \{demonstrating(man), white(paper), \\ fold(person,paper), fold(man, origami)\}\,.
    \end{gather*}
\item All facts that can be inferred from $K\!B$ and the predicted facts are used to augment the KG extracted from the video. Examples of such inferred facts are \textit{$change(person$, $paper)$} and \textit{$fold(male, origami)$}.\hfill$\lhd$
\end{enumerate}
\end{example}

\subsection{KG Extraction from Videos}
The above method can be used to extract a KG from any unstructured data $X$. For example, $X$ may contain sequences of word embeddings representing text, spectrograms representing audio, or image tensors representing still images. In each case, we only have to choose an encoder $f$, so that $f(x)$ is a fixed-length feature vector for each $x \in X$. We now describe how this method is applied when the input domain is videos.

%EG Instead of having this subsection, I would have a subsection called "Generalizing our model" (or something like that) where you discuss how bysimply changing the encoding you can basically extract knowledge from any type of data

Our encoder, $f$, is composed of a convolutional neural network, followed by a recurrent neural network.
Given a video $x = x_1,x_2,\dots,x_n$, we first process each video frame $x_i$ using a pre-trained frozen copy of VGG19~\cite{simonyan2014very} and take as $i$th frame encoding, $\zeta_i$, the output of the 3rd last layer. (This is a standard approach taken by others in the field; e.g., \cite{pan2017video,pan2016jointly,zhang2017task}.) The sequence of frame encodings $\zeta_1, \dots, \zeta_n$ is then passed through a gated recurrent unit (GRU) \cite{cho2014properties} to produce a sequence $\overline{\zeta}_1, \dots, \overline{\zeta}_n$. As a second stream, we also compute the feature vector from a frozen copy of the I3D network \cite{carreira2017quo}. The final output of the encoder is the concatenation of this I3D feature vector and a weighted sum of the $\overline{\zeta}_i$'s, weighted by a learnable $n$-dimensional vector $w$. Using a vector notation, with the VGG network applied along the first dimension, our encoder $f$ is defined by:
\begin{equation} \label{eq-encoder-def}
f(x) = w.GRU(VGG(x)) \cdot I3D(x)\,.
\end{equation}
The MLPs for each predicate have one hidden layer, as does the multi-classifier. The input size of the multi-classifier is the size of the video encoding, denoted $dim(f(x))$. The predicate-MLPs have input size $dim(f(x)) + D$ in the case of unary relations and $dim(f(x)) + 2D$ in the case of binary relations, where $D$ is the size of the individuals' vectors ($300$ in our case). Using the encoder defined in \eqref{eq-encoder-def}, $dim(f(x)) = 4096 + 1024 = 5120$. We restrict our attention to unary and binary predicates, but the framework can naturally be extended to include higher-order predicates.

%% file: text/results.tex
\section{Experimental Results} \label{sec-results}

In this section, we provide quantitative experimental results for the two datasets MSVD* and MSR-VTT*
and compare these results. We also report 
on ablation studies for our neural architecture and give some further qualitative results. 

\subsection{Quantitative Results} \label{subsec-quant-results}

Tables \ref{tab:MSVD*-results} and \ref{tab:MSRVTT*-results} display our performance on the two generated datasets MSVD* and \mbox{MSR-VTT*}, respectively, reporting both overall accuracy and F1-score. For each training example, these datasets contain many more negated facts than facts (see Section~\ref{sec-dataset} for details). This is the reason for such a large difference between accuracy and F1-score: the predictions are dominated by true negatives, which increase overall accuracy but have no effect on the F1-score. 
Accuracy and F1-score are harsh metrics, as they pertain to whole facts. An annotation would achieve 0 accuracy if it contained correct individuals and correct predicates, but related them in the wrong way, e.g., $fold(paper, person)$ or $carry(person,paper)$ in Fig.~\ref{fig:msrvtt-ex}. 
All results are taken from a held-out test set, using the train/val/test splits defined in the original MSVD and MSR-VTT datasets.

There are limited existing baselines (only \cite{Vasile}) for the task of KG extraction from videos. To benchmark our performance, we also compare to an artificial baseline composed of the video-captioning network from~\cite{olivastri2019end}, and the semantic parser used to create our datasets (described briefly in Section~\ref{sec-dataset}, and more in detail in the appendix).

\begin{table}[t] 
\centering
\caption{\small Results on the MSVD* video and derived KGs, as described in Section~\ref{sec-dataset}. The best results are in bold.} \label{tab:MSVD*-results}
    \medskip 
 \begin{tabular}{c|c|c|c|c|c|} 
\cline{2-5}
& \thead{F1-score} & \thead{Positive \\ Accuracy} & \thead{Negative \\ Accuracy} & \thead{Total \\ Accuracy}\\
\hline
\multicolumn{1}{|c|}{Our Model} & \textbf{13.99} & \textbf{12.65} & 99.20 & 22.16 \\ 
 \hline
\multicolumn{1}{|c|}{\begin{tabular}{@{}c@{}}Baseline\end{tabular}} & 13.5 & 7.53 & \textbf{99.96} & \textbf{25.55} \\
  \hline
\multicolumn{1}{|c|}{V\&L 2018}  & 6.11  & 3.36 & - & - \\
  \hline
 \end{tabular}\vspace*{2ex}
\end{table}

\begin{table}[t] 
\centering
\caption{\small Results on the MSR-VTT* video and derived KGs, as described in Section~\ref{sec-dataset}. The best results are in bold.} \label{tab:MSRVTT*-results}
    \medskip 
 \begin{tabular}{c|c|c|c|c|c|} 
\cline{2-5}
& \thead{F1-score} & \thead{Positive \\ Accuracy} & \thead{Negative \\ Accuracy} & \thead{Total \\Accuracy}\\
\hline
\multicolumn{1}{|c|}{Our Model} & 8.90 & \textbf{7.33} & 99.48 & 59.19 \\ 
 \hline
\multicolumn{1}{|c|}{\begin{tabular}{@{}c@{}} Baseline\end{tabular}} & \textbf{11.83} & 6.76 & \textbf{99.96} & \textbf{83.01} \\
  \hline
 \end{tabular}
\end{table}

Our model significantly outperforms the previous work by \cite{Vasile}, and performs on par with the video-captioning baseline. Importantly, it shows the best positive accuracy, which is the most difficult metric to score highly on.

%EG1 it would be nice to have a little table where the inference times are measured (you just need to take some and make an average)

%\subsection{Comparsion of Datasets}
The metrics, both for the baseline and for our network, are lower on MSR-VTT*. This is consistent with other reported results in the literature. For example \cite{olivastri2019end,xu2017learning,zhang2017task,chen2018less,wang2018reconstruction} all report lower video captioning performance on MSR-VTT than on MSVD. A likely reason is the greater vocabulary size of MSR-VTT: 29,316 vs.\ 13,010 for MSVD (figures taken from \cite{xu2016msr}). Although we exclude all but the most common individuals, a greater vocabulary size in the original NL captions would still make our task more difficult. It means that more facts will be excluded, and so there will be more annotations with missing information, which do not, therefore, fully describe the input~video.

\subsection{Ablation Studies}
To investigate the contribution of each part of our neural architecture, we perform ablation studies on the encoder $f$, predicate MLPs and the individual vector ($\{m_p \mid p\in P\}$ and $\{V_a \mid a \in A\}$ in the terminology of Section \ref{sec-method}).

The results for MSVD* and MSR-VTT* are shown in Tables~\ref{tab:msvd-ablation} and \ref{tab:msrvtt-ablation}, respectively. In the ``without encoder'' setting, the video encoding feature vector is replaced with a randomly generated vector. This tests whether the system is actually using information from the video or is just predicting common individuals and predicates, e.g., $talk(man)$. That the results in this setting are worse shows that information from the video is indeed being used. In the ``single MLP decoder'' setting, one MLP is used for all predicates, in contrast to the main system in which each predicate has its own MLP. This is to test whether predicate-specific information is being used, e.g., whether the system makes meaningfully different predictions for $throws(woman, ball)$ and $sees(woman,ball)$. Similarly, in the single-individual decoder setting, one vector is used for all individuals, in contrast to the main system in which each individual has its own vector. This tests if individual-specific information has been learned by the individual vectors. Both %of these 
decoder ablation settings produce worse results, which shows that the predicate-specific MLP and individual-specific vectors have learned some meaningful semantics of their respective predicates and individuals. Note that the results suffer more in these ablation settings on MSR-VTT* than on MSVD*. This is to be expected, as MSR-VTT* contains a larger vocabulary, which allows for more semantic differentiation between different predicates and between different individuals. 

\begin{table}[t]
 \centering
\caption{\small Ablation results for MSVD*. The best results are in bold.} \label{tab:msvd-ablation}
    \medskip 
 \begin{tabular}{c|c|c|c|c|c|} 
\cline{2-5}
& \thead{F1} & \thead{Positive \\ Accuracy} & \thead{Negative \\ Accuracy} & \thead{Total \\ Accuracy}\\
\hline
\multicolumn{1}{|c|}{Our Model} & \textbf{13.99} & \textbf{12.65} & 99.20 & 22.16 \\ 
 \hline
\multicolumn{1}{|c|}{Single MLP decoder} & 13.75 & 12.5 & 98.45 & 19.04 \\
  \hline
\multicolumn{1}{|c|}{\thead{Single individual \\ vector decoder}} & 10.72 & 9.91 & \textbf{99.32} & \textbf{22.50} \\
  \hline
\multicolumn{1}{|c|}{Without encoder} & 8.87 & 9.77 & 97.49 & 15.60 \\
  \hline
 \end{tabular}\vspace*{-2ex}
\end{table}

\begin{table}[t]
\centering
\caption{\small Ablation results for MSR-VTT*. The best results are in bold.} \label{tab:msrvtt-ablation}
    \medskip 
 \begin{tabular}{c|c|c|c|c|c|} 
\cline{2-5}
& \thead{F1} & \thead{Positive \\ Accuracy} & \thead{Negative \\ Accuracy} & \thead{Total \\ Accuracy}\\
\hline  
\multicolumn{1}{|c|}{Our Model} & \textbf{8.90} & \textbf{7.33} & \textbf{99.48} & \textbf{59.19} \\ 
 \hline
\multicolumn{1}{|c|}{Single MLP decoder} & 6.16 & 6.75 & 99.15 & 49.09 \\
  \hline
\multicolumn{1}{|c|}{\thead{Single individual \\ vector decoder}} & 3.84 & 3.94 & 99.51 & 58.72 \\
  \hline
\multicolumn{1}{|c|}{Without encoder} & 5.00 & 5.36 & 99.19 & 49.64 \\
  \hline
 \end{tabular}\vspace*{-2ex}
\end{table}

\subsection{Qualitative Results} \label{subsec-qual-results}
\begin{figure*}
\small
    \begin{tabular}{c l c l}
    \multirow{13}{*}{\includegraphics[width=0.2\textwidth]{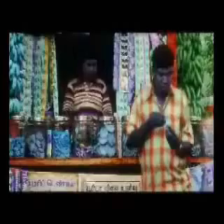}\hspace{-0.5ex}} &
    \textbf{(1) MSVD ground truth:} & \multirow{13}{*}{\includegraphics[width=0.2\textwidth]{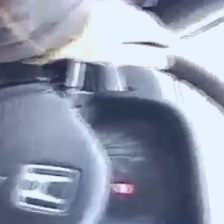}\hspace{-0.5ex}} & \textbf{(1) MSR-VTT ground truth:} \\ 
    &\text{a man is standing at the bus stop} &&\text{a man is driving a car}\\ [1ex] 
    &\textbf{(2) Predicted caption from~\cite{olivastri2019end}:} &&\textbf{(2) Predicted caption from~\cite{olivastri2019end}:}\\ 
    &\text{a man is talking on the phone}&&\text{a man is driving a car}\vspace*{-0.5ex}\\
    & \vspace*{-1.3ex} \\
    \cline{2-2}\cline{4-4}
    &  \vspace*{-0.8ex}\\
    &\textbf{(3) MSVD* ground truth:} &&\textbf{(3) MSR-VTT* ground truth:} \\ 
    &\textit{$stand(man)$} &&\text{$drive(man,car)$} \\
    &\textbf{(4) Predicted individuals:} &&\textbf{(4) Predicted individuals:} \\ 
    &\text{$man$}&&\text{$man$, $car$, $person$}  \\ 
    &\textbf{(5) Predicted facts:} &&\textbf{(5) Predicted facts:}\\ 
    &\text{$stand(man)$, $sit(man)$} && \text{$drive(man,car), drive(person,car)$}\\
    &\textbf{(6) Some inferred facts:} &&\textbf{(6) Some inferred facts:} \\  
    &\text{$stand(person)$, $stand(male)$} && \text{$drive(man,vehicle)$}%\vspace{3pt} \\
    %\textbf{(7) Parsed caption (baseline):} \\ 
    %\text{$talk(man)$}
    \end{tabular}
    
    \medskip \caption{%
        Left: the first frame from a video in the MSVD* test set, with the ground-truth caption, the caption produced by \cite{olivastri2019end}, the facts predicted by our system, and some further inferences made using $K\!B$ and these facts. Compared with a video captioning system, our output can be used to infer additional information about the video, such as the fact that it contains a male. Right: the second frame (first is very unclear) from a video in the MSR-VTT* test set, with the ground-truth caption, the caption produced by \cite{olivastri2019end}, the facts predicted by our system, and some further inferred facts. Although the video captioning model produces a sentence similar to the ground truth, it cannot be used for inference and so misses the additional inferred information.
    }
    \label{fig:msvd-msrvtt-ex}
\end{figure*}

To further evaluate the quality of our KG extractions, we present some manual inspections of images and the predicted facts. Fig.~\ref{fig:msvd-msrvtt-ex} shows the first frame from an MSVD* video with our predicted facts, and the same for a video from MSR-VTT*. (Fig.~\ref{fig:msrvtt-ex} shows another from MSR-VTT*.)

We can now infer additional facts from $K\!B$ and the facts generated by the video annotation network, which is one of the advantages of our approach of using KGs. The class $man$ is a subclass of $person$ and $male$ in WordNet, and we can apply this inference to all facts mentioning a man, producing $stand(male)$ and $stand(person)$ in Fig.~\ref{fig:msvd-msrvtt-ex}. If we wanted to determine how many videos in a database depicted at least one person, or how many depict males, or how many depict males sitting vs.\ standing, this would not be possible by merely annotating the videos with NL sentences. It is, on the other hand,  possible with our system, because of the inferences that can be made on top of a KG.
Finally, the qualitative examples show the limitations imposed by the smaller vocabulary size in MSVD*. 
As discussed in Section~\ref{sec-dataset}, we exclude individuals and relations that appear fewer than 50 times across the dataset. 
This means that there is sometimes insufficient material to fully describe a video. 
For example, the man in the video shown in Fig.~\ref{fig:msvd-msrvtt-ex} is eating a banana, and one of the MSVD captions expressed this. However, $banana$ appears fewer than 50 times, so it is excluded from our training data, and we cannot predict $eat(man,banana)$. (Recall from Section \ref{sec-dataset} that there are multiple annotations for each video, and we merge all of them in our dataset. The annotation shown in Fig.~\ref{fig:msvd-msrvtt-ex} is one that was not excluded.) 

%Note also, that despite producing a coherent and accurate natural language sentence, the video captioning system misses some \textit{direct} information present in the video, that is, information which could not even be inferred, such as the paper being white. Because the video captioning model is trained to mimic natural language sentences, there is a strong incentive to produce grammatical sentences. This encourages the production of short, simple sentences, a problem noted by \cite{yu2016video} among others, which can omit some information from the video. Our outputs are not bound by English grammar, and so are not biased towards short, simple annotations.

%% file: text/discussion.tex
\section{Further Discussion} \label{sec-discussion}
\vspace*{-0.5ex}
In this section, we further discuss the advantages of generating KGs compared to NL annotations with reference to the output of our model. We also examine our model as compared to the video-captioning baseline.  

\subsection{Value of KGs} \vspace*{-0.5ex}
One advantage of KGs is that they enable the application of automated data processing. An example is the application of inference rules to infer additional facts, which is illustrated by the results shown in Fig.~\ref{fig:msvd-msrvtt-ex}. The facts predicted by our model (3) allow the inference of further information (6), whereas the NL annotation produced by \cite{olivastri2019end} does not, even when it is exactly correct, as in the case of the \mbox{MSR-VTT} video. As well as producing additional information about the video contents, these inferences are especially useful for producing abstract classes and relations, such as $\mathit{male}$ in Fig.~\ref{fig:msvd-msrvtt-ex} or $\mathit{change}$ Fig.~\ref{fig:msrvtt-ex}. Other examples of automated data processing are query answering and search. If we are interested in videos that depict a male changing some object in some way, as happens in the video shown in Fig.~\ref{fig:msrvtt-ex}, then using the extracted KGs (including further inferences), we can simply return the videos where, for some individual $x$, the fact  $\mathit{change(male,x)}$ has been predicted. In contrast, NL annotations do not enable us to perform such a search. We cannot simply replace all occurrences of $\mathit{man}$, $\mathit{boy}$, $\mathit{guy}$ etc. with $\mathit{male}$, and all occurrences of $\mathit{fold}$, $\mathit{open}$, $\mathit{close}$, etc. with $\mathit{change}$ and then count the annotations that contain both $\mathit{male}$ and $\mathit{change}$. This would produce false positives with annotations like ``a~man is handed a folded sheet of paper''. In such a sentence, although $\mathit{male}$ and $\mathit{change}$ would both be present, they are not connected in the way that we are interested in. The sentence does not describe a scenario in which a male is changing something, even though it contains the words $\mathit{man}$ and $\mathit{fold}$. It is non-trivial to represent the information encoded in a NL sentence as a set of individuals and relations that hold between members of this set. To apply such a translation, we would need to employ a semantic parser, i.e., we would be using the method of the video-captioning baseline, and the demerits of this approach are discussed below. String matching in NL is not even able to detect the individuals that are present. The sentence ``a person folds a boy's school uniform'' could describe a video in which no boy is present. Our annotations represent information in the semantically structured format of a KG, from which information such as the above can be easily read off. Finally, we can see that the set of facts (equivalent to a KG) shown in Fig.~\ref{fig:msvd-msrvtt-ex} could be easily represented in a language other than English. The individuals and predicates in the vocabulary correspond to particular senses of particular words, rather than entire words, so disambiguating polysemous words is not required. To translate, e.g., the fact $\mathit{drive(man,car)}$ to German, we could simply look up the translation for the senses of each of these three words: drive $\to$ fahren, man $\to$ Mann, and car $\to$ Auto. Translating one item at a time, we then get $\mathit{fahren(Mann,Auto)}$. In contrast, translating the NL sentence would require both the disambiguation of polysemous words, and the extraction of English syntactic relations and grammatical inflections, and then the expression of these relations and inflections in German. 

\subsection{Comparison with Video-Captioning Baseline} \label{subsec:comp-with-baseline}
One possible method of producing KGs, and hence the advantages discussed above, is to first produce an NL annotation and then semantically parse this NL sentence to convert it to a set of facts. This is the method employed in the video-captioning baseline, to which we compare our performance in Section \ref{sec-results}. The quantitative comparison, across the multiple metrics and datasets, shows that it performs about equally well to our model. However, there are three reasons to believe that the latter is more promising for future progress. Firstly, the video-captioning baseline must solve two problems to produce an annotation: the problem of annotating videos with NL, and the problem of semantically parsing NL. This means it has two potential sources of error. However, errors in the second stage (that of semantic parsing) are not properly penalized in the results in Tables \ref{tab:MSVD*-results} and \ref{tab:MSRVTT*-results}, because the baseline uses the same semantic parser as was used to create the dataset. Therefore, if the video-captioning network correctly predicts the original NL caption, the result, after parsing, will automatically be correct according to our dataset. This will happen even if there was a mistake in the semantic parse. For example, in Fig.~\ref{fig:msrvtt-ex}, the NL annotation predicted by \cite{olivastri2019end} is ``a person is folding a piece of paper'',  which the parser then, incorrectly, parses as $\mathit{fold(person,piece)}$ (instead of $\mathit{fold(person,paper)}$). However, this same incorrect parse was made when creating the dataset, so the ground truth in \mbox{MSR-VTT*} is also $\mathit{fold(person,piece)}$, and the baseline is marked correct in this prediction. If there existed a dataset of videos with human-made KGs, the baseline would still require a semantic parser, and its accuracy would be limited by the parsing accuracy, whereas our model would not. 

A second advantage of our model is that we can restrict the vocabulary of our annotations when desired. We may, on occasion, only be interested in a certain set of individuals, and we could easily modify our network to only output annotations concerning these individuals. If, for example, we were only interested in determining whether a video contained any animals and what those animals were doing, we could remove all facts from our dataset other than those pertaining to animals and then train our model on the modified dataset. This would be a simpler task and be expected to give a higher accuracy. The video-captioning baseline, however, would still output an NL sentence as an intermediary step, and it would not be simple to apply the same restriction to this space of NL sentences. One may think that the vocabulary of the output language model in the video-captioning network could be restricted to only a certain, animal-related vocabulary. However, many words that have one meaning denoting an animal have also other meanings, e.g., ``fly'' or ``bear''. Again, this highlights the complexities of working with NL and so the undesirability of using it as an intermediary step to produce~a~KG.

Thirdly, the captioning and parsing approach is only possible where there already exist networks mapping the input domain to NL. This is true for videos, but need not be true in general. Our method, on the other hand, can be applied to any input domain, as long as a suitable encoder is used.

%% file: text/conclusion.tex
\section{Conclusion} \label{sec-conclusion}
This paper has proposed the task of KG extraction from videos, where a KG is composed of a set of facts that describes relations holding between individuals. We have provided arguments and empirical support for the advantages of KGs over NL annotations. No datasets currently exist of videos and corresponding KGs, so we have proposed a method to generate them from existing datasets of videos and NL annotations. Further, we have introduced a deep learning model for KG extraction, and evaluated its performance both qualitatively and quantitatively on two generated datasets.  

Future work includes exploring KG extraction from other input domains. Of particular interest is the application to text, where our model would perform a task akin to open information extraction. Another extension is to manually construct a dataset designed specifically for the purpose of KG extraction, rather than using those generated by our automated method.

%% file: text/appendix.tex
\section{Appendix}
To create the datasets MSVD* and MSR-VTT*, as described in Section \ref{sec-dataset}, we use a semantic parser to convert the natural language annotations into logical annotations. There are four steps to this conversion: parsing individual sentences, linking entities to the ontology (in our case, WordNet \cite{miller1990introduction}), merging the multiple annotations of each video and filtering out uncommon entities. They are described, respectively, in the 3 following subsections. 

\subsection{Parsing Natural Language Sentences}

First, the input natural language sentence is parsed using the Stanford parser. This identifies the part of speech of each word and constructs a grammatical dependency tree. In a grammatical dependency tree, a single word is identified as the root, and all other words are marked depending on their syntactic relation to their parent in the tree. The information produced by the Stanford Parser, for our purposes, can be thought of as marking each word in the input sentence with (i) a part of speech, (ii) another word in the sentence designated as the parent (unless this word is the root), and (iii) a grammatical dependency relation to the parent. An example of such a relation is `nsubj', which means this word occupies the subject position of its parent, its parent being a verb. Further discussion, including a list of all possible such dependencies, can be found in \cite{droganova2019towards, rademaker2019proceedings} and online at {https://universaldependencies.org/introduction.html}.
\begin{figure}[h]
    \centering
    \includegraphics[width=0.47\textwidth]{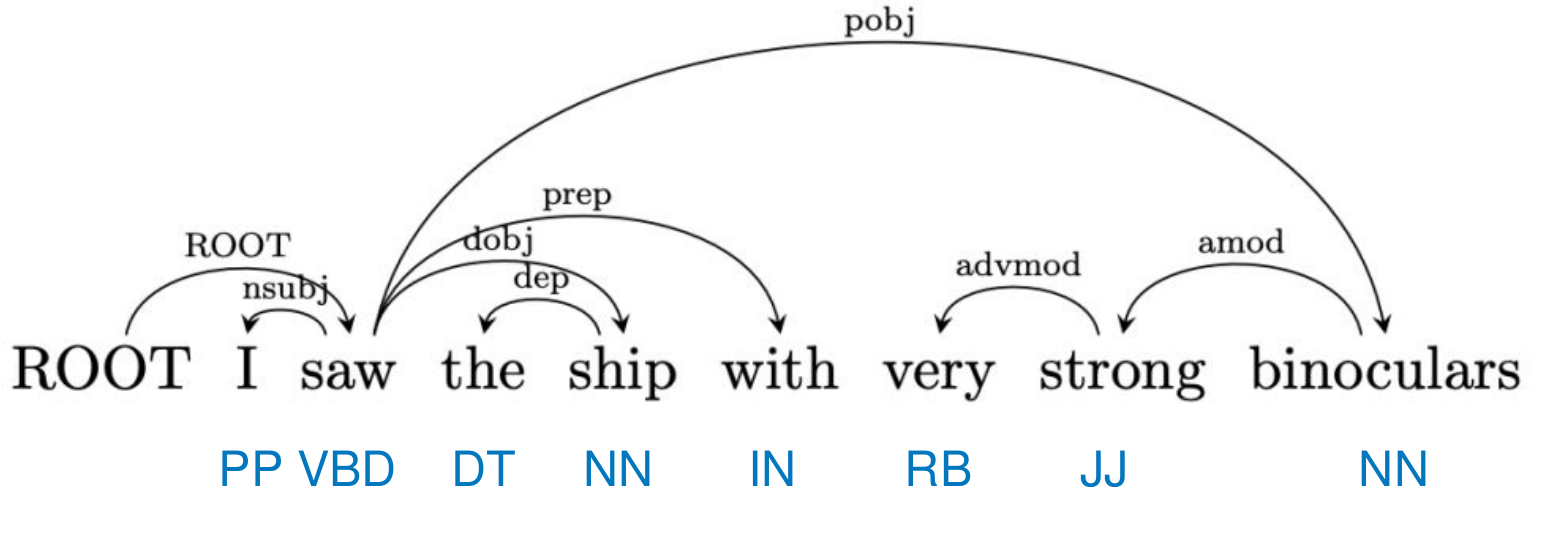}
    \caption{An example dependency parse of the sentence `I saw the ship with very strong binoculars'. The root is `saw', and all other words in the tree are marked for their syntactic relation to their parent. Parts of speech are shown in blue, using the Penn Treebank tagset, which is the tagset used by the Stanford parser. }
    \label{fig:my_label}
\end{figure}
From this syntactic parse, atoms are formed by Algorithm~\ref{alg:extract-all-atoms}.

\begin{algorithm}
\begin{algorithmic}[0]
\caption{ExtractRootAtom}
    \State \textbf{Inputs:} grammatical dependence tree of a natural language sentence
    \State \textbf{Output:} a single atom if one can be found, else None
    \State $s \gets$ the input natural language sentence 
    \State root $\gets$ the word designated as root
    \If {there is a word $w$ in $s$ with dependency marking `nsubj' to root}
        \State $subj \gets w$
    \Else
        \State Return None
    \EndIf
    \If{root is a verb}
         \If {there is a word $w'$ in $s$ with dependency marking `obj' to root}
            \State \Return $root(subj,w')$
        \Else
            \State \Return $root(subj)$
        \EndIf
    \EndIf
    \If {root is an adjective}
        \If {there is a word $c$ in $s$ with dependency marking `cop' to root}
            \State \Return $root(subj)$
        \Else  \State \Return null
        \EndIf
    \EndIf
    \If {root is an noun}
        \If {there is a word $c$ in $s$ with dependency marking `cop' to root and a word $p$ with dependency marking `case' to root and with part of speech `ADP'}
            \State \Return $p(arg1,root)$
        \EndIf
    \EndIf
          
\end{algorithmic}
\end{algorithm}
\begin{algorithm}
\begin{algorithmic}[0]
\caption{ExtractAllAtoms} \label{alg:extract-all-atoms}
    \State \textbf{Inputs:} a natural language sentence
    \State \textbf{Output:} a list of atoms
    \State $s \gets$ input natural language sentence 
    \State extractedAtoms $\gets []$
    \State $root \gets$ the word in sentence with dependency marking `root'
    \If {there is a word $w$ with dependency marking `obj' to root}
        \State append ExtractRootAtom($s$, $root$, $w$) to extractedAtoms
    \Else
        \State \Return null
    \EndIf
    \For {each word $conjWord$ in $s$ with dependency relation `cc'}
        \State $subClauseRoot$ $\gets$ parent of $conjWord$
        \State $subClauseSubj$ $\gets$ parent of $subClauseRoot$
        \State $subClauseAtom \gets$ ExtractRootAtom($s$,
        \\\qquad \qquad \qquad \qquad \qquad $subClauseRoot$,$subClauseSubj$)
        \State append $subClauseAtom$ to extractedAtoms
    \EndFor
    \For {each word $modWord$ in $s$  with dependency relation `amod' to to some other word $targetWord$}
        \State append $modWord(targetWord)$ to extractedAtoms
    \EndFor
    \For {each word $compWord$ in sentence with dependency relation `compound' to to some other word $targetWord$}
        \If {$compWord$ appears immediately before $targetWord$ in $s$}
            \State replace all instances of $targetWord$ in \\ \qquad \quad extractedAtoms with the concatenation \\
            \qquad \quad $compWord \cdot targetWord$
        \ElsIf {$compWord$ appears immediately before $targetWord$ in $s$}
            \State replace all instances of $targetWord$ in \\ \qquad \quad extractedAtoms with the concatenation \\
            \qquad \quad $targetWord \cdot compWord$  
        \EndIf
    \EndFor
    \State \Return extractedAtoms
\end{algorithmic}
\end{algorithm}
 
\subsection{Linking to an Ontology}
To apply inference rules from our ontology, we need to identify, for each predicate and object in our logical annotation, the corresponding entity from the ontology. This amounts to word-sense disambiguation. In WordNet, words are broken up, by sense, into different synsets. E.g., the word `bank' has one synset corresponding to the financial institution and another corresponding to the bank of a river. To identify the correct synset for each word $w$ in our logical annotations, we compute the doc2vec \cite{le2014distributed} representation of the original natural language sentence from which $w$ was taken, and the doc2vec representation of the definition of each WordNet synset containing the word $w$, and we pick the synset which has the most similar vector representation. Formally,
\[
syn = \underset{\{syn'|w \in syn'\}}{\mathrm{argmax}}doc2vec(def(syn'))doc2vec(s),
\]
where $def(syn')$ denotes the definition of synset $syn'$, and $s$ is the original natural language annotation sentence from which $w$ was taken. 

\subsection{Merging Logical Annotations}
A logical annotation is a set of atoms. We can merge all the logical annotations for each video by simply taking the union of the constituent atoms.

\subsection{Filtering the Vocabulary}
As described in Section \ref{sec-dataset}, we only consider ontology elements (i.e., objects and predicates) that appear a sufficient number of times in our dataset. Specifically, we remove all elements that appear in fewer than 50 data points, after the merging of captions in the previous step. Additionally, we exclude the verbs ``take'', ``do'', ``be'', and ``have''. These are semantically weak verbs, which normally function as copulas or other syntactic operators, and do not convey relevant information about the video. For example, the parse of the sentence ``a woman is standing at the top of the stairs'' should not include the atom $be(woman)$.